\documentclass[sigconf]{acmart}

\usepackage{booktabs}
\usepackage{siunitx}
\usepackage[font=small]{caption}

\setcopyright{acmlicensed}
\acmDOI{10.1145/3274895.3274927}
\acmISBN{978-1-4503-5889-7/18/11}
\acmConference[SIGSPATIAL '18]{26th ACM SIGSPATIAL International Conference on Advances in Geographic Information Systems}{November 6--9, 2018}{Seattle, WA, USA}
\acmYear{2018}
\copyrightyear{2018}
\acmArticle{4}
\acmPrice{15.00}
\acmBooktitle{26th ACM SIGSPATIAL International Conference on Advances in Geographic Information Systems (SIGSPATIAL '18), November 6--9, 2018, Seattle, WA, USA}

\ccsdesc[300]{Human-centered computing~Human computer interaction (HCI)}
\ccsdesc[300]{Applied computing~Cartography}

\keywords{Map Editing, Automatic Map Inference}

\begin{document}
\title{Machine-Assisted Map Editing}
\author{
	Favyen Bastani\textsuperscript{1}, Songtao He\textsuperscript{1}, Sofiane Abbar\textsuperscript{2}, Mohammad Alizadeh\textsuperscript{1}, \\
	Hari Balakrishnan\textsuperscript{1}, Sanjay Chawla\textsuperscript{2}, Sam Madden\textsuperscript{1} \\
	\textsuperscript{1}MIT CSAIL, \textsuperscript{2}Qatar Computing Research Institute, HBKU \\
	\textsuperscript{1}\{fbastani,songtao,alizadeh,hari,madden\}@csail.mit.edu, \textsuperscript{2}\{sabbar,schawla\}@hbku.edu.qa
}

\renewcommand{\shortauthors}{Favyen Bastani et al.}
\newcommand{\authorsxyz}{Favyen Bastani, Songtao He, Sofiane Abbar, Mohammad Alizadeh, Hari Balakrishnan, Sanjay Chawla, Sam Madden}

\begin{abstract}
Mapping road networks today is labor-intensive. As a result, road maps have poor coverage outside urban centers in many countries. Systems to automatically infer road network graphs from aerial imagery and GPS trajectories have been proposed to improve coverage of road maps. However, because of high error rates, these systems have not been adopted by mapping communities. We propose \emph{machine-assisted map editing}, where automatic map inference is integrated into existing, human-centric map editing workflows. To realize this, we build Machine-Assisted iD (MAiD), where we extend the web-based OpenStreetMap editor, iD, with machine-assistance functionality. We complement MAiD with a novel approach for inferring road topology from aerial imagery that combines the speed of prior segmentation approaches with the accuracy of prior iterative graph construction methods. We design MAiD to tackle the addition of major, arterial roads in regions where existing maps have poor coverage, and the incremental improvement of coverage in regions where major roads are already mapped. We conduct two user studies and find that, when participants are given a fixed time to map roads, they are able to add as much as 3.5x more roads with MAiD.
\end{abstract}

\maketitle

\section{Introduction}

In many countries, road maps have poor coverage outside urban centers. For example, in Indonesia, roads in the OpenStreetMap dataset~\cite{openstreetmap} cover only 55\% of the country's road infrastructure\footnote{55\% coverage is computed by summing length of OSM roads and comparing against The World Factbook known road network length (https://www.mapbox.com/data-platform/country/\#indonesia).}; the closest mapped road to a small village may be tens of miles away.
Map coverage improves slowly because mapping road networks is very labor-intensive. For example, when adding roads visible in aerial imagery, users need to perform repeated clicks to draw lines corresponding to road segments.

This issue has motivated significant interest in \emph{automatic map inference}. Several systems have been proposed for automatically constructing road maps from aerial imagery~\cite{cheng2017automatic, deeproadmapper} and GPS trajectories~\cite{biagioni2012map, kharita}. Yet, despite over a decade of research in this space, these systems have not gained traction in OpenStreetMap and other mapping communities. Indeed, OpenStreetMap contributors continue to add roads solely by tracing them by hand.

Fundamentally, high error rates make full automation impractical. Even state-of-the-art automatic map inference approaches have error rates between 5\% and 10\%~\cite{roadtracer, kharita}. Navigating the road network using road maps with such high frequencies of errors would be virtually impossible.

Thus, we believe that automatic map inference can only be useful when it is integrated with existing, human-centric map editing workflows. In this paper, we propose \emph{machine-assisted map editing} to do exactly that.

Our primary contribution is the design and development of Machine-Assisted iD (MAiD), where we integrate machine-assistance functionality into iD, a web-based OpenStreetMap editor. At its core, MAiD replaces manual tracing of roads with human validation of automatically inferred road segments. We designed MAiD with a holistic view of the map editing process, focusing on the parts of the workflow that can benefit substantially from machine-assistance. Specifically, MAiD accelerates map editing in two ways.

In regions where the map has low coverage, MAiD focuses the user's effort on validation of major, arterial roads that form the backbone of the road network. Incorporating these roads into the map is very useful since arterial roads are crucial to many routes. At the same time, because major roads span large distances, validating automatically inferred segments covering major roads is significantly faster than tracing the roads manually. However, road networks inferred by map inference methods include both major and minor roads. Thus, we propose a novel shortest-path-based pruning scheme that operates on an inferred road network graph to retain only inferred segments that correspond to major roads.

In regions where the map has high coverage, further improving map coverage requires users to painstakingly scan the aerial imagery and other data sources for unmapped roads. We reduce this scanning time by adding a ``teleport'' feature that immediately pans the user to an inferred road segment. Because many inferred segments correspond to service roads and residential roads that are not crucial to the road network, we design a segment ranking scheme to prioritize segments that are more useful.

We find that existing schemes to automatically infer roads from aerial imagery are not suitable for the interactive workflow in MAiD. Segmentation-based approaches~\cite{cheng2017automatic, qian2017road, deeproadmapper}, which apply a CNN to label pixels in the imagery as ``road'' or ``non-road'', have low accuracy because they require an error-prone post-processing stage to extract a road network graph from the pixel labels. Iterative graph construction (IGC) approaches~\cite{roadtracer, ventura2017iterative} improve accuracy by extracting road topology directly from the CNN, but have execution times six times slower than segmentation, which is too slow for interactivity.

To facilitate machine-assisted interactive mapping, we develop a novel method for extracting road topology from aerial imagery that combines the speed of segmentation-based approaches with the high-accuracy of iterative graph construction (IGC) approaches.
Our method adapts the IGC process to use a CNN that outputs road directions for all pixels in one shot; this substantially reduces the number of CNN evaluations, thereby reducing inference time for IGC by almost 8x with near-identical accuracy. Furthermore, in contrast to prior work, our approach infers not only unmapped roads, but also their connections to an existing road network graph.

To evaluate MAiD, we conduct two user studies where we compare the mapping productivity of our validation-based editor (coupled with our map inference approach) to an editor that requires manual tracing. In the first study, we ask participants to map roads in an area of Indonesia with no coverage in OpenStreetMap, with the goal of maximizing the percentage of houses covered by the mapped road network. We find that, given a fixed time to map roads, participants are able to produce road network graphs with 1.7x the coverage and comparable error when using MAiD. In the second study, participants add roads in an area of Washington where major roads are already mapped. With MAiD, participants add 3.5x more roads with comparable error.

In summary, the contributions of this paper are:

\begin{itemize}
    \item We develop MAiD, a machine-assisted map editing tool that enables efficient human validation of automatically inferred roads.
    \item We propose a novel pruning algorithm and teleport feature that focus validation efforts on tasks where machine-assisted editing offers the greatest improvement in mapping productivity.
    \item We develop an approach for inferring road topology from aerial imagery that complements MAiD by improving on prior work.
    \item We conduct user studies to evaluate MAiD in realistic editing scenarios, where we use the current state of OpenStreetMap, and find that MAiD improves mapping productivity by as much as 3.5x.
\end{itemize}

The remainder of this paper is organized as follows. In Section \ref{sec:related}, we discuss related work. Then, in Section \ref{sec:maid}, we detail the machine-assisted map editing features that we develop to incorporate automatic map inference into the map editing process. In Section \ref{sec:improving}, we introduce our novel approach for map inference from aerial imagery. Finally, we evaluate MAiD and our map inference algorithm in Section \ref{sec:evaluation}, and conclude in Section \ref{sec:conclusion}.

\section{Related Work} \label{sec:related}

\noindent \textbf{Inference from Aerial Imagery.} Most state-of-the-art approaches for inferring road maps from aerial imagery apply convolutional neural networks (CNNs) to segment the imagery for ``road'' and ``non-road'' pixels, and then post-process the segmentation output to extract a road network graph. Cheng et al. develop a cascaded CNN architecture with two jointly trained components, where the first component detects pixels on the road, and the second focuses on pixels close to the road centerline~\cite{cheng2017automatic}. They then threshold and thin the centerline segmentation output to extract a graph.

Shi et al. propose improving the segmentation output by using a conditional generative adversarial network~\cite{qian2017road}. They train the segmentation CNN not only to output the ground truth labels (with a mean-squared-error loss), but also to fool a discriminator CNN that is trained to distinguish between the ground truth labels and the segmentation CNN outputs.

DeepRoadMapper adds an additional post-processing step to infer missing connections in the initial extracted road network~\cite{deeproadmapper}. Candidate missing connections are generated by performing a shortest path search on a graph defined by the segmentation probabilities. Then, a separate CNN is trained to identify correct missing connections.

Rather than segmenting the imagery, RoadTracer~\cite{roadtracer} and IDL~\cite{ventura2017iterative} employ an iterative graph construction (IGC) approach that extracts roads via a series of steps in a search process. On each step, a CNN is queried to determine what direction to move in the search, and a road segment is added to a partial road network graph in that direction. Although IGC methods improve accuracy, they are an order of magnitude slower in execution time than segmentation approaches and thus not suitable in interactive settings.

\smallskip
\noindent \textbf{Inference from GPS Trajectories.} Instead of using aerial imagery, several systems propose to instead infer maps from GPS trajectories~\cite{biagioni2012map, kharita, buchin2017clustering}. Each GPS trajectory is a sequence of GPS positions observed as a vehicle moves along the road network during one trip. Kernel density estimation, clustering, and trajectory merging algorithms can be applied on datasets of trajectories collected across many vehicles and trips to produce a road map.

Automatically extending existing maps using GPS trajectory data has also been studied. Most of these approaches share a common map-matching-based architecture: they attempt to match trajectories to the road network, and portions of trajectories that fail to match are clustered to produce new road segments~\cite{crowdatlas, cobweb, hymu}. MapFuse instead proposes a more general map fusion scheme, where the road network inferred by any map inference approach is fused with the existing map~\cite{mapfuse}.

\section{Machine-Assisted \lowercase{i}D} \label{sec:maid}

\begin{figure*}
\begin{center}
	\includegraphics[width=\linewidth]{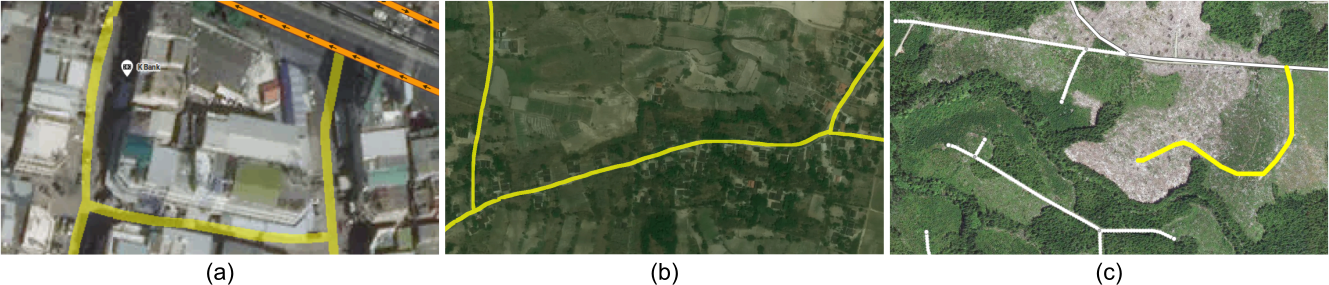}
\end{center}
	\caption{Tracing the short, straight roads in (a) by hand can be done as quickly as validating inferred segments (yellow) corresponding to such roads. But, tracing the long, arterial roads in (b) is much more tedious. In (c), because most roads are already covered by the map (white segments), finding unmapped roads (yellow) is time-consuming.}
\label{fig:ondemand}
\end{figure*}

\begin{figure*}
\begin{center}
	\includegraphics[width=\linewidth]{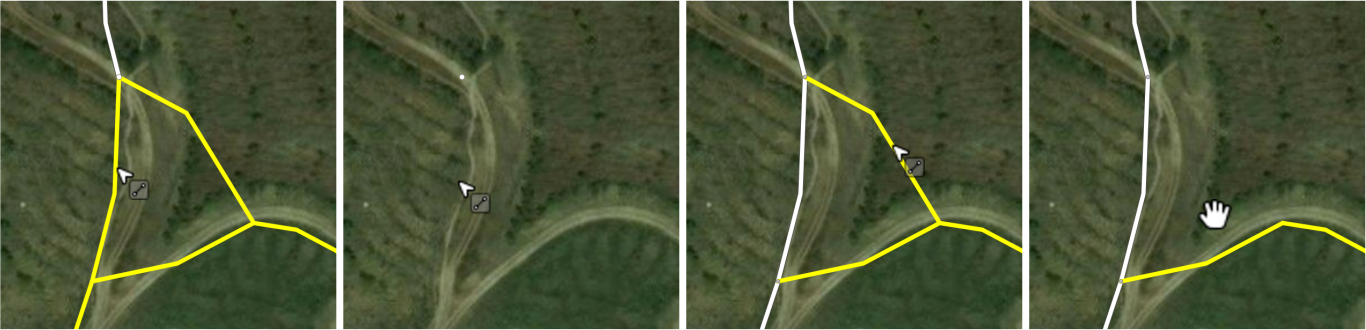}
\end{center}
	\caption{The MAiD editing workflow. On the left, there is one mapped road segment in white, and several inferred roads in yellow. Then, the user hides the overlay to verify the position of the roads in the imagery. After clicking on two of the yellow segments, these segments are added to the map. Finally, on the right, the user removes an incorrect inferred segment by right-clicking.}
\label{fig:uiseries}
\end{figure*}

Building machine-assistance into the map editor to maximize the improvement in mapping productivity is not straightforward. The obvious approach, which we implemented in an early prototype, would have users draw rectangles around areas with missing roads; the system would then run automatic map inference to infer segments covering those roads, and create an overlay containing these segments. However, we found that often, most of the inferred segments corresponded to straight, short service and residential roads that were not much faster to validate than to trace by hand. For example, when tracing the unmapped roads in Figure \ref{fig:ondemand}(a), users spend most of the time looking at the imagery and identifying the positions of roads. The actual tracing can be done quickly -- since the roads are straight and cover a small area, only a few clicks are needed. Because users still need to examine the imagery when validating the inferred segments, this system does not reduce mapping time.

Thus, we instead focus on map editing in two specific contexts where machine-assistance can improve mapping productivity substantially. In regions where the map has low coverage, tracing major, arterial roads like those in Figure \ref{fig:ondemand}(b) is tedious because the roads span large distances; thus, if we can focus validation on major roads, machine-assistance can significantly speed up mapping. On the other hand, in regions like Figure \ref{fig:ondemand}(c) where the map has high coverage, further improving coverage requires users to painstakingly scan the imagery for unmapped roads. We develop a teleport feature in the map editor that eliminates this time-consuming process by panning users immediately to groups of unmapped roads.

In Section \ref{sec:ui}, we first introduce the user interface that we design to incorporate validation of inferred segments into an existing map editor. Then, in Section \ref{sec:pruning}, we detail our pruning scheme that retains inferred segments covering major roads, and in Section \ref{sec:teleport}, we describe our teleport functionality.

\subsection{UI for Validation} \label{sec:ui}

We build MAiD, where we incorporate our machine-assistance features into iD, a web-based OpenStreetMap editor.

A road network graph is a graph where vertices are annotated with spatial coordinates (latitude and longitude) and edges correspond to straight-line road segments. MAiD inputs an existing road network graph $G_0 = (V_0, E_0)$ containing roads already incorporated in the map. To use MAiD, users first select a region of interest for improving map coverage. MAiD runs an automatic map inference approach in this region to obtain an inferred road network graph $G = (V, E)$  containing inferred segments corresponding to unmapped roads. $G$ should satisfy $E_0 \cap E = \varnothing$; however, $G$ and $G_0$ share vertices at the points where inferred segments connect with the existing map.

To make validation of automatically inferred segments intuitive, MAiD then produces a yellow overlay that highlights inferred segments in $G$ over the aerial imagery. Although the overlay is partially transparent, in some cases it is nevertheless difficult to verify the position of the road in the imagery when the overlay is active; thus, users can press and hold a key to temporarily hide the overlay so that they can consult the imagery.

After verifying that an inferred segment is correct, users can left-click the segment to incorporate it into the map. Existing functionality in the editor can then be used to adjust the geometry or topology of the road. If an inferred segment is erroneous, users can either ignore the segment, or right-click on the segment to hide it.

Figure \ref{fig:uiseries} shows the MAiD editing workflow.

\subsection{Mapping Major Roads} \label{sec:pruning}

However, we find that this validation-based UI alone does not significantly increase mapping productivity. To address this, we first consider adding roads in regions where the map has low coverage. In practice, when mapping these regions, users typically focus on tracing major, arterial roads that form the backbone of the road network. More precisely, major roads connect centers of activity within a city, or link towns and villages outside cities; in OpenStreetMap, these roads are labelled ``primary'', ``secondary'', or ``tertiary''. Users skip short, minor roads because they are not useful until these important links are mapped. Because major roads span large distances, though, tracing them is slow. Thus, validation can substantially reduce the mapping time for these roads.

Supporting efficient validation of major roads requires the pruning of inferred segments corresponding to minor roads. However, automatically distinguishing major roads is difficult. Often, major roads have the same width and appearance as minor roads in aerial imagery. Similarly, while major roads in general have higher coverage by GPS trajectories, more trips may traverse minor roads in population centers than major roads in rural regions.

Rather than detecting major roads from the data source, we propose a shortest-path-based pruning scheme that operates on an inferred road network graph to retain only inferred segments that correspond to major roads. Intuitively, major roads are related to shortest paths: because major roads offer fast connections between far apart locations, they should appear on shortest paths between such locations.

We initially applied betweenness centrality~\cite{betweenness}, a measure of edge importance based on shortest paths. The betweenness centrality of an edge is the number of shortest paths between unique origin-destination pairs that pass through the edge. (When computing shortest paths in the road network graph, the length of an edge is simply the distance between its endpoints.) Formally, for a road network graph $G = (V, E)$, the betweenness centrality of an edge $e$ is:

$$g(e) = \sum\limits_{s \neq t \in V} I[e \in \text{shortest-path}(s, t)]$$

We can then filter edges in the graph by thresholding based on the betweenness centrality scores.

\begin{figure}
\begin{center}
	\includegraphics[width=0.7\linewidth]{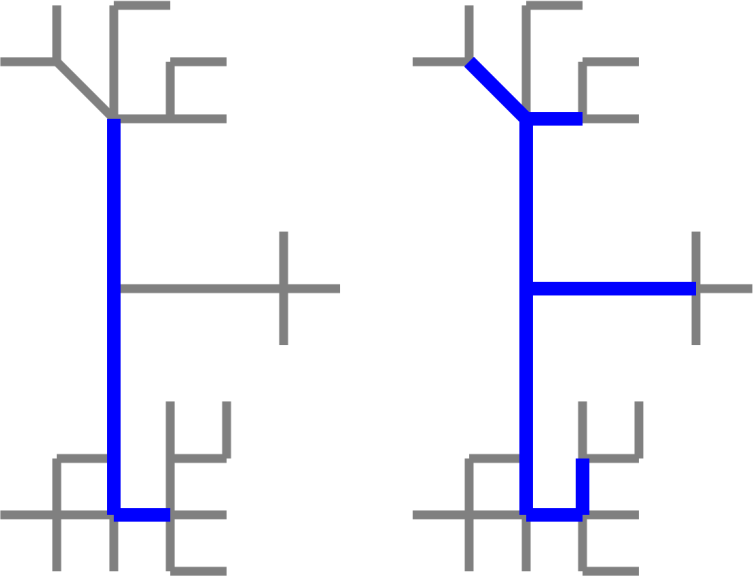}
\end{center}
	\caption{Thresholding by betweenness centrality performs poorly. Grey segments are pruned to produce a road network graph containing the blue segments. On the left, a high threshold misses the road to the eastern cluster. On the right, a low threshold includes small roads in the northern and southern clusters.}
\label{fig:bc_problem}
\end{figure}

However, we find that segments with high betweenness centrality often do not correspond to important links in the road network. When using a high threshold, the segments produced after thresholding cover major roads connecting dense clusters in the original graph, but miss connections to smaller clusters. When using a low threshold, most major roads are retained, but minor roads in dense clusters are also retained. Figure \ref{fig:bc_problem} shows an example of this issue. Additionally, different regions require very different thresholds.

Thus, we propose an adaptation of betweenness centrality for our pruning problem.

\smallskip
\noindent \textbf{Pruning Minor Roads} Fundamentally, betweenness centrality fails to consider the overall spatial distribution of vertices in the road network graph. Dense but compact clusters in the road network should not have an undue influence on the pruning process.

Our pruning approach builds on our earlier intuition, that major roads connect far apart locations. Thus, rather than considering all shortest paths in the graph, we focus on long shortest paths. Additionally, we observe that the path may use minor roads near the source and near the destination, but edges on the middle of a shortest path are more likely to be major roads.

We first cluster the vertices of the road network. Then, we compute shortest paths between cluster centers that are at least a minimum radius $R$ apart. Rather than computing a score and then thresholding on the score, we build a set of edges $E_{\text{major}}$ containing edges corresponding to major roads that we will retain. For each shortest path, we trim a fixed distance from the ends of the path, and add all edges in the remaining middle of the path to $E_{\text{major}}$. We prune any edge that does not appear in $E_{\text{major}}$.

\begin{figure}
\begin{center}
	\includegraphics[width=0.7\linewidth]{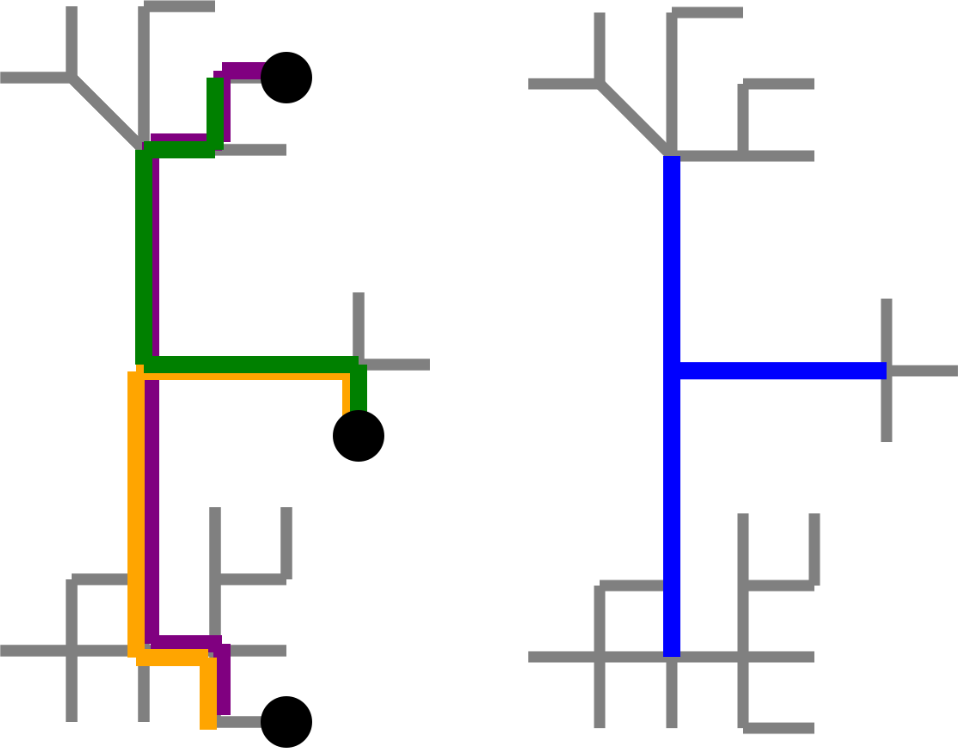}
\end{center}
	\caption{Our pruning approach. We cluster the graph, and compute shortest paths between cluster centers (left). Edges are trimmed from the beginning and end of these paths, and the remaining edges from the paths are retained (right).}
\label{fig:pruning}
\end{figure}

Figure \ref{fig:pruning} illustrates our approach.

We find that our approach is robust to the choice of the clustering algorithm. Clustering is primarily used to avoid placing cluster centers at vertices that are at the end of a long road that only connects a small number of destinations (and, thus, isn't a major road). In our implementation, we use a simple grid-based clustering scheme: we divide the road network into a grid of $r \times r$ cells, remove cells that contain less than a minimum number of vertices, and then place cluster centers at the mean position of vertices in the remaining cells. We use $r = 1$ km, $R = 5$ km.

In practice, we find that for constant $R$, the runtime of our approach scales linearly with the length of the input road network.

\begin{figure}
\begin{center}
	\includegraphics[width=\linewidth]{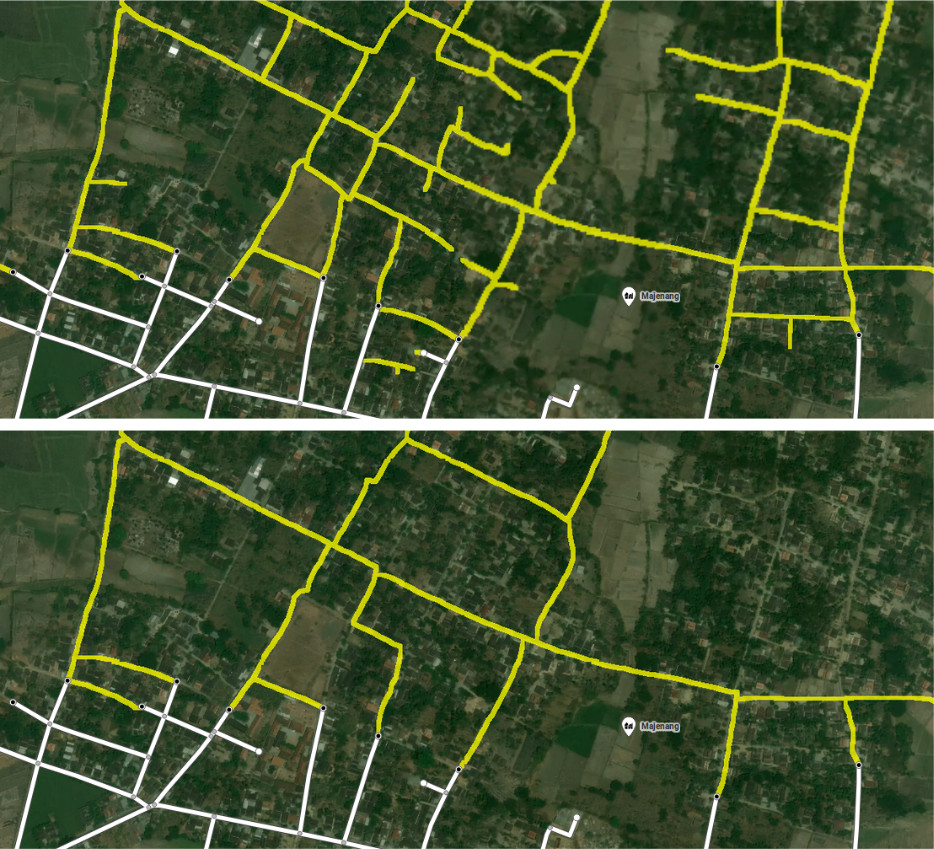}
\end{center}
	\caption{Pruning in MAiD. Above, the overlay includes all automatically inferred road segments. Below, pruning is applied to retain only segments corresponding to major roads.}
\label{fig:maidpruning}
\end{figure}

\smallskip
\noindent \textbf{MAiD Implementation.} We add a button to toggle between an overlay containing all inferred roads, and an overlay after pruning. Figure \ref{fig:maidpruning} shows an example of pruning in Indonesia.

\subsection{Teleporting to Unmapped Roads} \label{sec:teleport}

In regions where the map already has high coverage, further improving the map coverage is tedious. Because most roads already appear in the map, users need to slowly scan the aerial imagery to identify unmapped roads in a very time-consuming process.

To address this, we add a teleport capability into the map editor, which pans the editor viewport directly to an area with unmapped roads. Specifically, we identify connected components in the inferred road network $G$, and pan to a connected component. This functionality enables a user to teleport to an unmapped component, add the roads, and then immediately teleport to another component. By eliminating the time cost of searching for unmapped roads in the imagery, we speed up the mapping process significantly.

However, there may be hundreds of thousands of connected components, and validating all of the components may not be practical. Thus, we propose a prioritization scheme so that longer roads that offer more alternate connections between points on the existing road network are validated first.

Let $\text{area}(C)$ be the area of a convex hull containing the edges of a connected component $C$ in $G$, and let $\text{conn}(C)$ be the number of vertices that appear in both $C$ and $G_0$, i.e., the number of connections between the existing road network and the inferred component $C$. We rank connected components by $score(C) = \text{area}(C) + \lambda \text{conn}(C)$, for a weighting factor $\lambda$.

\section{Fast, Accurate Map Inference} \label{sec:improving}

In the map inference problem, given an existing road network graph $G_0 = (V_0, E_0)$, we want to produce an inferred road network graph $G = (V, E)$ where each edge in $E$ corresponds to a road segment visible in the imagery but missing from the existing map.

Prior work in extracting road topology from aerial imagery generally employ a two-stage segmentation-based architecture. First, a convolutional neural network (CNN) is trained to label pixels in the aerial imagery as either ``road'' or ``non-road''. To extract a road network graph, the CNN output is passed through a heuristic post-processing pipeline that begins with thresholding, morphological thinning~\cite{thinning}, and Douglas-Peucker simplification~\cite{douglaspeucker}. However, robustly extracting a graph from the CNN output is challenging, and the post-processing pipeline is error-prone; often, noise in the CNN output is amplified in the final road network graph~\cite{roadtracer}.

Rather than segmenting the imagery, RoadTracer~\cite{roadtracer} and IDL~\cite{ventura2017iterative} propose an iterative graph construction (IGC) approach that improves accuracy by deriving the road network graph more directly from the CNN. IGC uses a step-by-step process to construct the graph, where each step contributes a short segment of road to a partial graph. To decide where to place this segment, IGC queries the CNN, which outputs the most likely direction of an unexplored road. Because we query the CNN on each step, though, IGC requires an order of magnitude more inference steps than segmentation-based approaches. We find that IGC is over six times slower than segmentation.

Thus, existing map inference methods are not suitable for the interactive nature of MAiD.

\begin{figure}
\begin{center}
	\includegraphics[width=\linewidth]{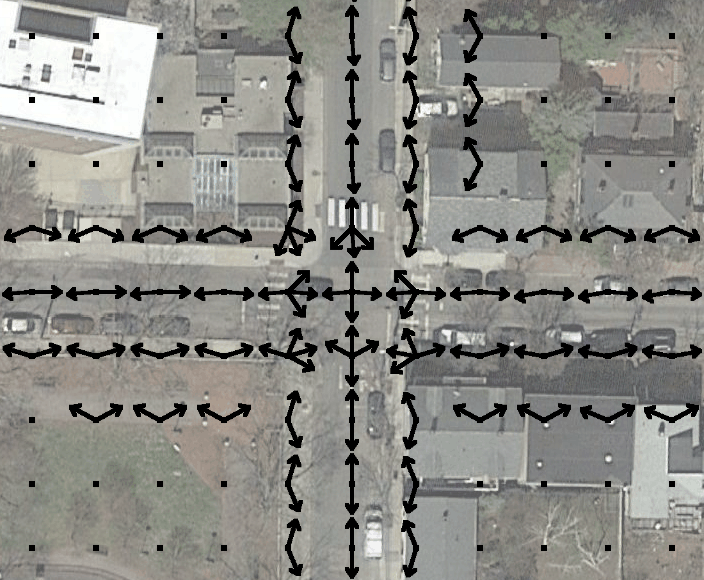}
\end{center}
	\caption{We train a CNN to annotate pixels with directions of roads indicated by the arrows.}
\label{fig:flow}
\end{figure}

We combine the two-stage architecture of segmentation-based approaches with the road-direction output and iterative search process of IGC to achieve a high-speed, high-accuracy approach. In the first stage, rather than labeling pixels as road or non-road, we apply a CNN on the aerial imagery to annotate each pixel in the imagery with the \emph{direction} of roads near that pixel. Figure \ref{fig:flow} shows an example of these annotations. In the second stage, we iteratively construct a road network graph by following these directions in a search process.

\smallskip
\noindent \textbf{Ground Truth Direction Labels} We first describe how we obtain the per-pixel road-direction information shown in Figure \ref{fig:flow} from a ground truth road network $G^* = (V^*, E^*)$. For each pixel $(i, j)$, we compute a set of angles $A^*_{i, j}$. If there are no edges in $G^*$ within a matching threshold of $(i, j)$, $A^*_{i, j} = \varnothing$.

Otherwise, suppose $e$ is the closest edge to $(i, j)$, and let $p$ be the closest point on $e$ computed by projecting $(i, j)$ onto $e$. Let $P_{i, j}$ be the set of points in $G^*$ that are a fixed distance $D$ from $p$; put another way, $P_{i, j}$ contains each point $p'$ such that $p'$ falls on some edge $e' \in E^*$, and the shortest distance from $p$ to $p'$ in $G^*$ is $D$.

Then, $A^*_{i, j} = \{\text{angle}(p' - (i, j)) \mid p' \in P_{i, j}\}$, i.e., $A^*_{i,  j}$ contains the angle from $(i, j)$ to each point in $P_{i, j}$.

\begin{figure}
\begin{center}
	\includegraphics[width=\linewidth]{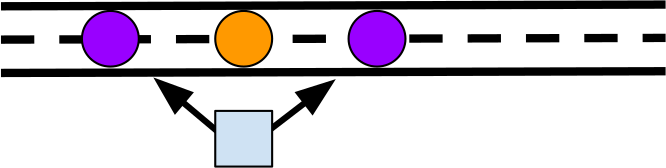}
\end{center}
	\caption{We compute $A^*$ at the position indicated by the blue square. We project the position onto the closest road to obtain the orange point. The purple points are exactly $D$ distance along the road network from the orange point. $A^*$ at this position contains angles from the blue square to the purple points.}
\label{fig:road_directions}
\end{figure}

Figure \ref{fig:road_directions} shows an example of computing $A^*_{i, j}$.

\smallskip
\noindent \textbf{Representing Road Directions.} We represent $A^*$ as a 3-dimensional matrix $U^*$ that can be output by a CNN. We discretize the space of angles corresponding to road directions into $b = 64$ buckets, where the $k$th bucket covers the range of angles from $\frac{2k\pi}{b}$ to $\frac{2(k+1)\pi}{b}$. We then convert each set of road directions $A^*_{i, j}$ to a $b$-vector $u^*(i, j)$, where $u^*(i, j)_k = 1$ if there is some angle in $A^*_{i, j}$ falling into the $k$th angle bucket, and $u^*(i, j)_k = 0$ otherwise. Then, $U^*_{i, j, k} = u(i, j)_k$.

\smallskip
\noindent \textbf{CNN Architecture.} Our CNN model inputs the RGB channels from the $w \times h$ aerial imagery, and outputs a $w \times h \times b$ matrix $U$.

We apply 16 convolutional layers in a U-Net-like configuration~\cite{unet}, where the first 11 layers downsample to 1/32 the input resolution, and the last 5 layers upsample back up to 1/4 the input resolution. We use $3 \times 3$ kernels in all layers. We use sigmoid activation in the output layer, and rectified linear activation in all other layers. We use batch normalization in the 14 intermediate layers between the input and output layers.

We train the CNN on random $256 \times 256$ crops of the imagery with a mean-squared-error loss, $\sum_{i, j, k} (U_{i, j, k} - U^*_{i, j, k})^2$, and use the ADAM gradient descent optimizer~\cite{adam}.

\smallskip
\noindent \textbf{Search Process.} At inference time, after applying the CNN on aerial imagery to obtain $U$, we perform a search process using the predicted road directions in $U$ to derive a road network graph. We adapt the search process from IGC. Essentially, the search iteratively follows directions in $U$ to construct the graph, adding a fixed-length road segment on each step.

We assume that a set of points $V_{\text{init}}$ known to be on the road network are provided. If there is an existing map $G_0$, we will show later how to derive $V_{\text{init}}$ from $G_0$. Otherwise, $V_{\text{init}}$ may be derived from peaks in the two-dimensional matrix $m(U)_{i, j} = \max_k U_{i, j, k}$. We initialize a road network graph $G$ and a vertex stack $S$, and populate both with vertices at the points in $V_{\text{init}}$.

Let $S_{\text{top}}$ be the vertex at the head of $S$, and let $u_{\text{top}} = U(S_{\text{top}})$ be the vector in $U$ corresponding to the position of $S_{\text{top}}$. For an angle bucket $a$, $u_{\text{top},a}$ is the predicted likelihood that there is a road in the direction corresponding to $a$ from $S_{\text{top}}$. On each step of the search, we use $u_{\text{top}}$ to decide whether there is a road segment adjacent to $S_{\text{top}}$ that hasn't yet been mapped in $G$, and if there is such a segment, what direction that segment extends in.

We first mask out directions in $u_{\text{top}}$ corresponding to roads already incorporated into $G$ to obtain a masked vector $\text{mask}(u_{\text{top}})$; we will discuss the masking procedure later. Masking ensures that we do not add a road segment that duplicates a road that we captured earlier in the search process. Then, $\text{mask}(u_{\text{top}})_a$ is the likelihood that there is an \emph{unexplored} road in the direction $a$.

If the maximum likelihood after masking, $\max_a \text{mask}(u_{\text{top}})_a$, exceeds a threshold $T$, then we decide to add a road segment. Let $a_{\text{best}} = \text{argmax}_a \text{mask}(u_{\text{top}})_a$ be the direction with highest likelihood after masking, and let $w_{a_\text{best}}$ be a unit-vector corresponding to angle bucket $a_{\text{best}}$. We add a vertex $v$ at $S_{\text{top}} + D w_{a_{\text{best}}}$, i.e., at the point $D$ away from $S_{\text{top}}$ in the direction indicated by $a_{\text{best}}$. We then add an edge $(S_{\text{top}}, v)$, and push $v$ onto $S$.

Otherwise, if $\max_a \text{mask}(u_{\text{top}})_a < T$, we stop searching from $S_{\text{top}}$ (since there are no unexplored directions with a high enough confidence in $U$) by popping $S_{\text{top}}$ from $S$. On the next search step, we will return to the previous vertex in $S$.

\begin{figure}
\begin{center}
	\includegraphics[width=\linewidth]{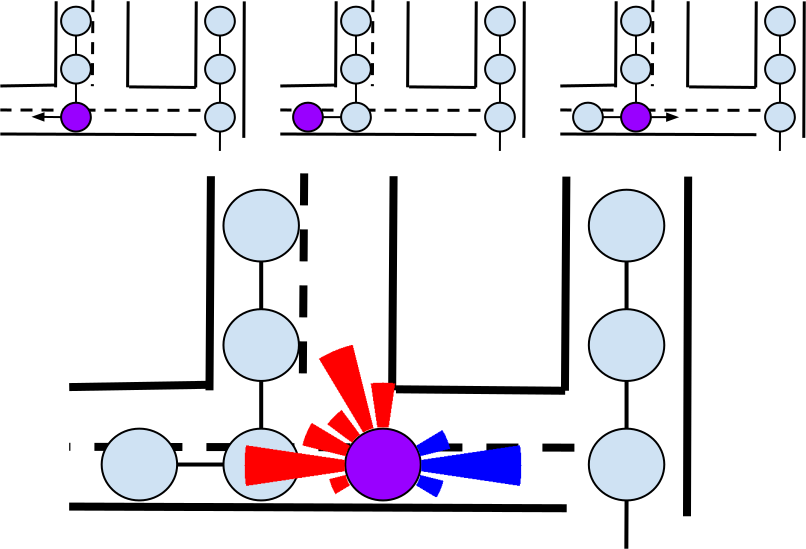}
\end{center}
	\caption{The search process. $S_{\text{top}}$ is purple, and other vertices in $G$ are blue. At top, we show three iterations that expand $G$. At bottom, we show the decision process on the fourth iteration in detail, with red indicating masked out directions and blue indicating remaining directions.}
\label{fig:search}
\end{figure}

Figure \ref{fig:search} illustrates the search process. At the top, we show three search iterations, where we add a segment, stop, and then add another segment. At the bottom, we show the fourth iteration in detail. Likelihoods in $u_{\text{top}}$ peak to the left, topleft, and right. After masking, only the blue bars pointing right remain, since the left and topleft directions correspond to roads that we already mapped. We take the maximum of these remaining likelihoods and compare to the threshold $T$ to decide whether to add a segment from $S_{\text{top}}$ or stop.

When searching, we may need to merge the current search path with other parts of the graph. For example, in the fourth iteration of Figure \ref{fig:search}, we approach an intersection on the right where the perpendicular road was already added to $G$ earlier in the search. We handle merging with a simple heuristic that avoids creating spurious loops. Let $N_k(S_{\text{top}})$ be the set of vertices within $k$ edges from $S_{\text{top}}$. If $S_{\text{top}}$ is within $2D$ of another vertex $v$ in $G$ such that $v \not\in N_5(S_{\text{top}})$, then we add an edge $(S_{\text{top}}, v)$.

\smallskip
\noindent \textbf{Masking Explored Roads.} If we do not mask during the search, then we would repeatedly explore the same road in a loop. Masking out directions corresponding to roads that were explored earlier in the search ensures that roads are not duplicated in $G$.

We first mask out directions that are similar to the angle of edges incident to $S_{\text{top}}$. For each edge $e$ incident to $S_{\text{top}}$, if the angle of $e$ falls in bucket $a$, we set $\text{mask}(u_{\text{top}})_{a+k} = 0 \, \forall k, -5 \le k \le 5$.

However, this is not sufficient. In the fourth iteration of Figure \ref{fig:search}, there is an explored road to the north of $S_{\text{top}}$, but that road is connected to a neighbor west of $S_{\text{top}}$ rather than directly to $S_{\text{top}}$. Thus, we also mask directions that are similar to the angle from $S_{\text{top}}$ to any vertex in $N_5(S_{\text{top}})$.

\smallskip
\noindent \textbf{Extending an Existing Map.} We now show how to apply our map inference approach to improve an existing road network graph $G_0$. Our key insight is that we can use points on $G_0$ as starting locations for the search process. Then, when new road segments are inferred, these points inform the connectivity between the new segments and $G_0$.

We first preprocess $G_0$ to derive a densified existing map $G'_0$. Densification is necessary because there may not be a vertex at the point where an unmapped road branches off from a road in the existing map.
To densify $G_0 = (V_0, E_0)$, for each $e \in E_0$ where $\text{length}(e) > D$, we add $\lfloor \frac{\text{length}(e)}{D} \rfloor$ evenly spaced vertices between the endpoints of $e$, and replace $e$ with edges between those vertices. This densification preprocessing produces a base map $G'_0$ where the distance between adjacent vertices is at most $D$.

To initialize the search, we set $G = G'_0$, and add vertices in $G'_0$ to $S$. We then run the search process to termination. The search produces a merged road network graph $G$ that contains both segments in the existing map and inferred segments. We extract the inferred road network graph by removing the edges of $G'_0$ from this output graph $G$.

\section{Evaluation} \label{sec:evaluation}

To evaluate MAiD, we perform two user studies. In Section \ref{sec:eval_study1}, we consider a region of Indonesia where OpenStreetMap has poor coverage to evaluate our pruning approach. In Section \ref{sec:eval_study2}, we turn to a region of Washington where major roads are already mapped to evaluate the teleport functionality.

In Section \ref{sec:eval_infer}, we compare our map inference scheme against prior work in map inference from aerial imagery on the RoadTracer dataset~\cite{roadtracer}. We show qualitative results when using MAiD with our map inference approach in Section \ref{sec:eval_qualitative}.

\subsection{Indonesia Region: Low Coverage} \label{sec:eval_study1}

We first conduct a user study to evaluate mapping productivity when adding roads in a small area of Indonesia with no coverage in OSM. With MAiD, the interface includes a yellow overlay of automatically inferred roads; to obtain these roads, we generate an inferred graph from aerial imagery using our map inference method, and then apply our pruning algorithm to retain only the major roads. After validating the geometry of a road, the user can click it to incorporate the road into the map. In the baseline unmodified editor, users manually trace roads by performing repeated clicks along the road in the imagery.

\smallskip
\noindent \textbf{Procedure.} The task is to map major roads in a region using the imagery, with the goal of maximizing coverage in terms of the percentage of houses within 1000 ft of the road network. Users are also asked to produce a connected road network, and to minimize the distance between road segments and the road position in the imagery. We define two metrics to measure this distance: road geometry error (RGE), the average distance between road segments that the participants add and a ground truth map that we hand label, and max-RGE, the maximum distance.

Ten volunteers, all graduate and postdoctoral students age 20-30, participate in our study. We use a within-subjects design; five participants perform the task first on the baseline editor, and then on MAiD, and five participants proceed in the opposite order.

Participants perform the experiment in a twenty-minute session. We select three regions from the unmapped area: an example region, a training region, and a test region. We first introduce participants to the iD editor, and enumerate the editor features as they add one road. We then describe the task, and show them the example region where the task has already been completed. Participants briefly practice the task on the training region, and then have four minutes to perform the task on a test region. We repeat the training and testing for both editors.

We choose the test region so that it is too large to map within the allotted four minutes. We then evaluate the road network graphs that the participants produce using each editor in terms of coverage (percentage of houses covered), RGE, and max-RGE.

\begin{figure}
\begin{center}
	\includegraphics[width=0.8\linewidth]{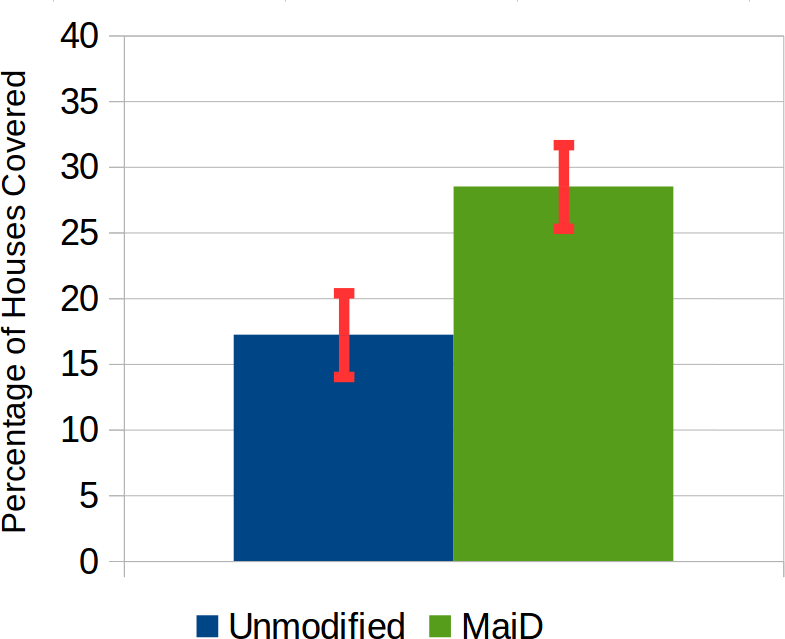}
\end{center}
	\caption{Mean and standard error of the percentage of houses covered in maps produced with the baseline editor and MAiD.}
\label{fig:study1_coverage}
\end{figure}

\smallskip
\noindent \textbf{Results.} We report the mean and standard error of the percentage of houses covered by the participants with the two editors in Figure \ref{fig:study1_coverage}. We find that MAiD improves the mean percentage covered by 1.7x (from 17\% to 29\%). While manually tracing a major road may take 15-30 clicks, the road can be captured with one click in MAiD after the geometry of an inferred segment is verified.

RGE and max-RGE are comparable for both editors, although there is more variance between participants with the baseline editor because the roads are manually traced. The mean and standard error of RGE across the participants is $\SI{5.0}{\meter} \pm \SI{0.6}{\meter}$ with the baseline, and $\SI{4.1}{\meter} \pm \SI{0.1}{\meter}$ with MAiD. For max-RGE, it is $\SI{33}{\meter} \pm \SI{5}{\meter}$ with the baseline, and $\SI{20}{\meter} \pm \SI{1}{\meter}$ with MAiD.

\subsection{Washington Region: High Coverage} \label{sec:eval_study2}

Next, we evaluate mapping productivity in a high-coverage region of rural Washington. With MAiD, users can press a Teleport button to immediately pan to a group of unmapped roads. A yellow overlay includes all inferred segments covering those roads; we do not use our pruning approach for this study. With the baseline editor, users need to pan around the imagery to find unmapped roads. After finding an unmapped road, users manually trace it.

\smallskip
\noindent \textbf{Procedure.} The task is to add roads that are visible in the aerial imagery but not yet covered by the map. Because major roads in this region are already mapped, rather than measuring house coverage, we ask users to add as much length of unmapped roads as possible. We again ask users to minimize the distance between road segments and the road position in the imagery, and to ensure that new segments are connected to the existing map.

Ten volunteers (consisting of graduate students, postdoctoral students, and professional software engineers all age 20-30) participate in our study. We again use a within-subjects design and counterbalance the order of the baseline editor and MAiD.

Participants perform the experiment in a fifteen-to-twenty minute session. For each editing interface, we first provide instructions on the task and editor functionality (accompanied by a 30-second video where we use the editor), and show images of example unmapped roads. Participants then practice the task on a training region in a warm-up phase, with a suggested duration of two to three minutes. After participants finish the warm-up, they are given three minutes to perform the task on a test region. As before, we repeat training and testing for both interfaces.

We evaluate the road network graphs that the participants produce in terms of total road length, RGE, and max-RGE.

\begin{figure}
\begin{center}
	\includegraphics[width=0.9\linewidth]{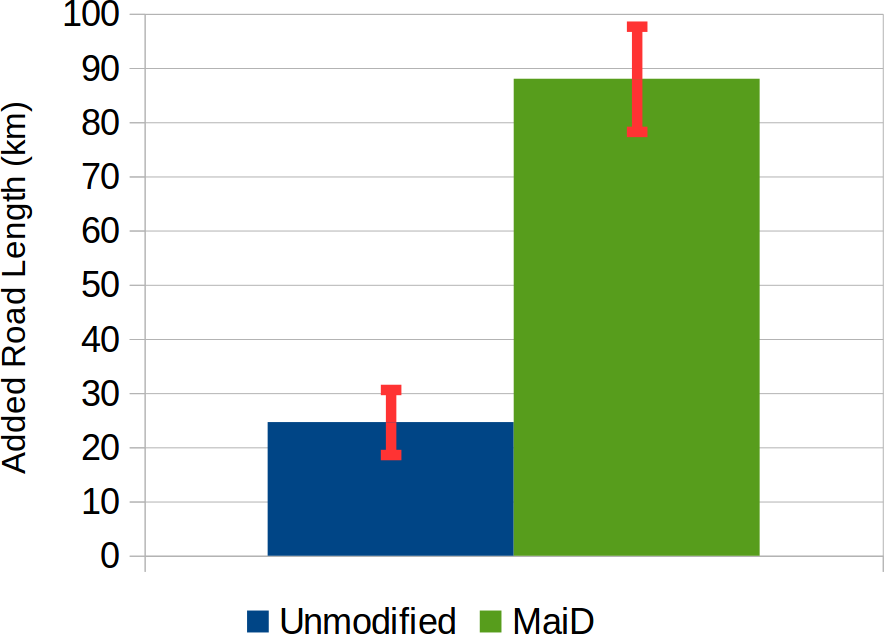}
\end{center}
	\caption{Mean and standard error of the total length of added roads with the baseline editor and MAiD.}
\label{fig:study2_length}
\end{figure}

\smallskip
\noindent \textbf{Results.} We report the mean and standard error of total road length added by the participants in Figure \ref{fig:study2_length}. MAiD improves mapping productivity in terms of road length by 3.5x (from 25 km to 88 km). Most of this improvement can be attributed to the teleport functionality eliminating the need for panning around the imagery to find unmapped roads. Additionally, though, because teleport prioritizes large unmapped components with many connections to the existing road network, validating these components is much faster than manually tracing them.

As before, RGE and max-RGE are comparable for the two editors. Mean and standard error of RGE is $\SI{7.0}{\meter} \pm \SI{0.7}{\meter}$ with the baseline editor, and $\SI{5.3}{\meter} \pm \SI{0.1}{\meter}$ with MAiD. For max-RGE, it is $\SI{53}{\meter} \pm \SI{14}{\meter}$ with the baseline, and $\SI{39}{\meter} \pm \SI{4}{\meter}$ with MAiD.

\subsection{Automatic Map Inference} \label{sec:eval_infer}

\noindent \textbf{Dataset.} We evaluate our approach for inferring road topology from aerial imagery on the RoadTracer dataset \cite{roadtracer}, which contains imagery and ground truth road network graphs from forty cities. The data is split into a training set and a test set; the test set includes data for a 16 sq km region around the city centers of 15 cities, while the training set contains data from 25 other cities. Imagery is from Google Maps, and road network data is from OpenStreetMap.

The test set includes 9 cities in the U.S., 3 in Canada, and 1 in each of France, the Netherlands, and Japan.

\smallskip
\noindent \textbf{Baselines.} We compare against the baseline segmentation approach and the IGC implementation from \cite{roadtracer}. The segmentation approach applies a 13-layer CNN, and then extracts a road network graph using thresholding, thinning, and refinement. The IGC approach, RoadTracer, trains a CNN using a supervised dynamic labels procedure that resembles reinforcement learning. This approach achieves state-of-the-art performance on the dataset, on which DeepRoadMapper \cite{deeproadmapper} has also been evaluated.

\smallskip
\noindent \textbf{Metrics.} We evaluate the road network graphs output by the map inference schemes on the TOPO metric~\cite{topo}, which is commonly used in the automatic road map inference literature~\cite{ahmed2015comparison}. TOPO evaluates both the geometrical accuracy (how closely the inferred segments align with the actual road) and the topological accuracy (correct connectivity) of an inferred map. It simulates an agent traveling on the road network from an origin location, and compares the destinations that can be reached within a fixed radius in the inferred map with those that can be reached in the ground truth map. This comparison is repeated over a large number of randomly selected origins to obtain an average precision and recall.

We also evaluate the execution time of the schemes on an AWS p2.xlarge instance with an NVIDIA Tesla K80 GPU.

\begin{figure}
\begin{center}
	\includegraphics[width=\linewidth]{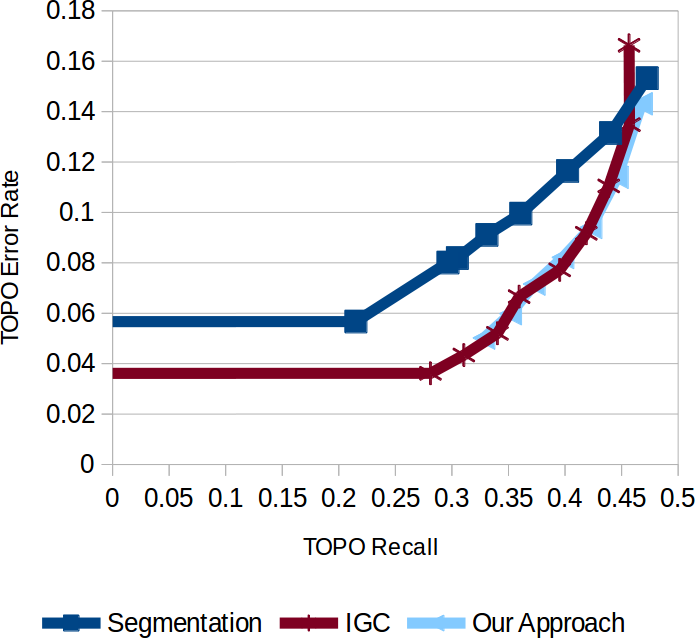}
\end{center}
	\caption{Average TOPO recall and error rate ($1 - \text{precision}$) over the 15 test cities. For each approach, we show a precision-recall curve over different parameter choices.}
\label{fig:topo_infer}
\end{figure}

\begin{table}
    \centering
    \begin{tabular}{| c | r | r | r |}
        \hline
        Approach & Inference & Processing & Total \\
        \hline
        Segmentation & 84 sec & 74 sec & 158 sec\\
        \hline
        IGC & 947 sec & 116 sec & 1063 sec \\
        \hline
        Our Approach & 85 sec & 51 sec & 136 sec \\
        \hline
    \end{tabular}
    \caption{Execution time evaluation. Inference is time spent applying the CNN, while processing includes all other execution time.}
    \label{tab:time_infer}
\end{table}

\smallskip
\noindent \textbf{Results.} We show TOPO precision-recall curves obtained by varying parameter choices in Figure \ref{fig:topo_infer}, and average execution time in the 15-square-km test regions for parameters that correspond to a 10\% error rate in Table \ref{tab:time_infer}. We find that our approach exhibits both the high-accuracy of IGC and the speed of segmentation methods.

Our map inference approach has comparable TOPO performance to IGC, while outperforming the segmentation approach on error rate by up to 1.6x. This improvement in error rate is crucial for machine-assisted map editing as it reduces the time users spend validating incorrect inferred segments.

On execution time, our approach performs comparably to the segmentation approach, while IGC is almost 8x slower. A low execution time is crucial to MAiD's interactive workflow. Users can explore a new region for two to three minutes while the automatic map inference approach runs; however, a fifteen-minute runtime breaks the workflow.

\subsection{Qualitative Results} \label{sec:eval_qualitative}

\begin{figure*}
\begin{center}
	\includegraphics[width=\linewidth]{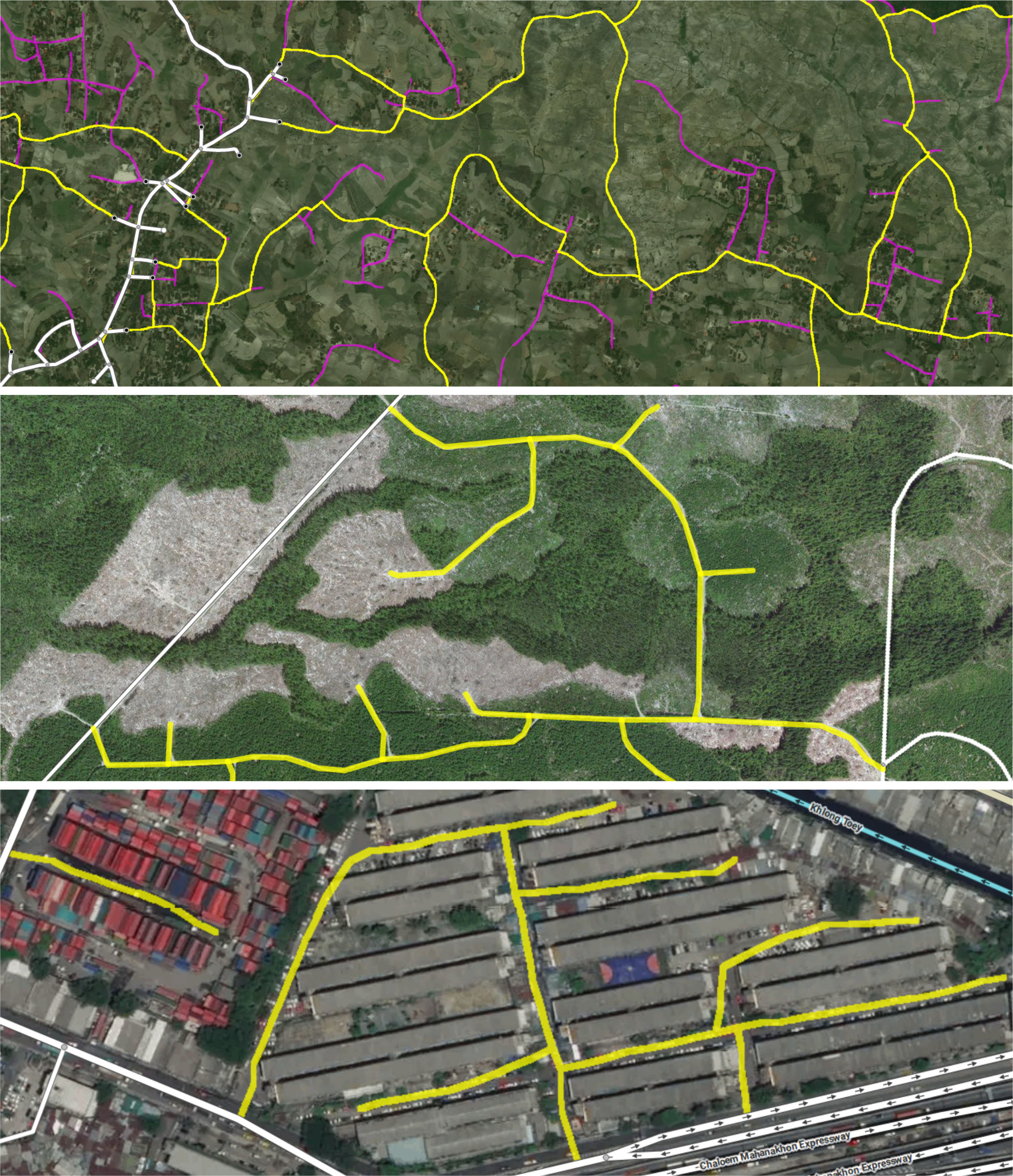}
\end{center}
	\caption{Qualitative results from MAiD with our map inference algorithm. Segments in the existing map are in white. We show our pruning approach applied on a region of Indonesia in the top image, with pruned roads in purple and retained roads in yellow. The middle and bottom images show connected components of inferred segments that the teleport feature pans the user to, in Washington and Bangkok respectively.}
\label{fig:qualitative}
\end{figure*}

In Figure \ref{fig:qualitative}, we show qualitative results from MAiD when using segments inferred by our map inference algorithm.

\section{Conclusion} \label{sec:conclusion}

Full automation for building road maps has proven unfeasible due to high error rates in automatic map inference methods. We instead propose machine-assisted map editing, where we integrate automatically inferred road segments into the existing map editing process by having humans validate these segments before the segments are incorporated into the map. Our map editor, Machine-Assisted iD (MAiD), improves mapping productivity by as much as 3.5x by focusing on tasks where machine-assistance provides the most benefit. We believe that by improving mapping productivity, MAiD has the potential to substantially improve coverage in road maps.

\bibliographystyle{ACM-Reference-Format}
\bibliography{paper}

\end{document}